\documentclass[runningheads]{llncs}
\usepackage{graphicx}
\graphicspath{{./Fig/}}
\DeclareGraphicsExtensions{.pdf,.jpg,.png}

\usepackage[caption=false,font=footnotesize]{subfig}

\usepackage[hidelinks]{hyperref} 
\hypersetup{colorlinks,
citecolor=black,%
filecolor=black,%
linkcolor=black,%
urlcolor=blue
}
\usepackage{doi}
\usepackage[english]{babel}
\usepackage[strings]{underscore} 

\usepackage[utf8]{inputenc}
\usepackage[T1]{fontenc}

\usepackage{color} 

\usepackage{amsfonts}
\usepackage{amsmath,bm}

\usepackage{mathrsfs}
\usepackage{scalerel}

\usepackage{array}

\usepackage{url}

\usepackage{stfloats}

\usepackage{paralist}

\usepackage{etoolbox}
\newcommand{\zerodisplayskips}{%
  \setlength{\abovedisplayskip}{1pt}%
  \setlength{\belowdisplayskip}{1pt}%
  \setlength{\abovedisplayshortskip}{1pt}%
  \setlength{\belowdisplayshortskip}{1pt}}
\appto{\normalsize}{\zerodisplayskips}
\appto{\small}{\zerodisplayskips}
\appto{\footnotesize}{\zerodisplayskips}

\BeforeBeginEnvironment{figure}{\vskip-3ex}
\AfterEndEnvironment{figure}{\vskip-3ex}

\newcommand\scale[2]{\vstretch{#1}{\hstretch{#1}{#2}}}
	
\newcommand{\LP}{\if@draft
		\mathbin{\ooalign{$\bigtriangleup$\crcr\hidewidth \raise.14em\hbox{$\scale{0.7}{\scriptscriptstyle+}$}\hidewidth}}
	\else
		\mathbin{\mathpalette\LIPcls+}
	\fi}
\newcommand{\LM}{\if@draft
		\mathbin{\ooalign{$\bigtriangleup$\crcr\hidewidth \raise.14em\hbox{$\scale{0.7}{\scriptscriptstyle-}$}\hidewidth}}
	\else
		\mathbin{\mathpalette\LIPcls-}
	\fi}
\newcommand{\LT}{\if@draft
	  \mathbin{\ooalign{$\bigtriangleup$\crcr\hidewidth \raise.14em\hbox{$\scale{0.7}{\scriptscriptstyle\times}$}\hidewidth}}
	\else
		\mathbin{\mathpalette\LIPcls\times}
	\fi}

\newcommand{\LIPcls}[2]{%
  \ooalign{$#1\bigtriangleup$\crcr  \hidewidth\raisefix{#1}\hbox{$#1\scale{0.45}{\bm{#2}}$}\hidewidth}}

\def\raisefix#1{%
  \ifx#1\displaystyle
    \raise.14em
  \else
    \ifx#1\textstyle
      \raise.14em
    \else
      \ifx#1\scriptstyle
        \raise.112em
      \else
        \raise.0933em
      \fi
    \fi
  \fi
}

\newcommand{\LIPplus}{\LP}
\newcommand{\LIPminus}{\LM}
\newcommand{\LIPtimes}{\LT}

\newcommand{\Real}{\mathbb R}
\newcommand{\Realb}{\overline{\Real}}

\newcommand{\la}{\lambda}

\newcommand{\I}{\mathcal{I}}
\newcommand{\Ib}{\overline{\I}}

\newcommand{\Fcurv}{\mathcal{F}}
\newcommand{\Fcurvb}{\overline{\Fcurv}}

\newif\ifcompile
\compiletrue 

\newcommand{\myinclude}[1]{\if@draft
	\include{#1}
\else
  \input{#1}
\fi}

\begin{document}
\title{A link between the multiplicative and additive functional Asplund's metrics}
%
\titlerunning{Linking the multiplicative and additive Asplund's metrics}
%
\author{Guillaume Noyel\inst{1,2}\orcidID{0000-0002-7374-548X}}
\authorrunning{G. Noyel}
%
\institute{University of Strathclyde Institute of Global Public Health, Dardilly, Lyon, France \and
International Prevention Research Institute, Dardilly, Lyon, France\\
\email{guillaume.noyel@i-pri.org}}
\maketitle              
\begin{abstract}
Functional Asplund's metrics were recently introduced to perform pattern matching robust to lighting changes thanks to double-sided probing in the Logarithmic Image Processing (LIP) framework. Two metrics were defined, namely the LIP-multiplicative Asplund's metric which is robust to variations of object thickness (or opacity) and the LIP-additive Asplund's metric which is robust to variations of camera exposure-time (or light intensity). Maps of distances - i.e. maps of these metric values - were also computed between a reference template and an image. Recently, it was proven that the map of LIP-multiplicative Asplund's distances corresponds to mathematical morphology operations. In this paper, the link between both metrics and between their corresponding distance maps will be demonstrated. It will be shown that the map of LIP-additive Asplund's distances of an image can be computed from the map of the LIP-multiplicative Asplund's distance of a transform of this image and vice-versa. Both maps will be related by the LIP isomorphism which will allow to pass from the image space of the LIP-additive distance map to the positive real function space of the LIP-multiplicative distance map. Experiments will illustrate this relation and the robustness of the LIP-additive Asplund's metric to lighting changes.

\end{abstract}

%
%
\section{Introduction}
\label{sec:intro}

Functional Asplund's metrics defined in the Logarithmic Image Processing (LIP) framework \cite{Jourlin2016} are useful tools to compare images. They possess the interesting property of \textit{insensitivity to lighting changes}. Two functional metrics were introduced by Jourlin et al:
\begin{inparaenum}[(i)] 
\item firstly, the LIP-multiplicative Asplund's metric which is based on the LIP-multiplicative law is robust to a variation of object opacity (or thickness) \cite{Jourlin2012,Jourlin2014} and
\item secondly, the LIP-additive Asplund's metric which is defined with the LIP-additive law is robust to a variation of the light intensity (or the camera exposure-time) \cite{Jourlin2016}.
\end{inparaenum}
They both extend to grey level images the Asplund's metric for binary shapes \cite{Asplund1960,Grunbaum1963} which is insensitive to an homothety (or a magnification) of the shapes. In each functional metric, one image is selected as a probe and is compared to the other one on its two sides after a LIP-multiplication by a scalar (i.e. an homothety) or a LIP-addition of a constant (i.e. an intensity translation). Maps of Asplund's distances can also be computed between the neighbourhood of each pixel and a probe function defined on a sliding window \cite{Jourlin2012}. Noyel and Jourlin have shown \cite{Noyel2017a,Noyel2017b} that the map of LIP-multiplicative of Asplund's distance is a combination of Mathematical Morphology (MM) operations \cite{Matheron1967,Serra1982,Serra1988,Heijmans1994,Najman2013}. 

In the literature, other approaches of double-sided probing were defined. E.g., in the hit-or-miss transform \cite{Serra1988} and in its extension to grey level images \cite{Khosravi1996}, a unique structuring element was matched with the image from above and from below. In \cite{Banon1997},  Banon et al. translated a unique template two times along the grey level axis. An erosion and an anti-dilation were used to count the pixels whose values were in between the two translated templates.
In an approach inspired by the computation of the Hausdorff distance, Odone et al. \cite{Odone2001} checked if the grey values of an image were included in an ``interval'' around the other. The interval was obtained by a 3D dilation of the other image and was vertically translated for each point of the first image. If a sufficient number of the image points were in the ``interval'', then the template was considered as matched.
Barat et al. \cite{Barat2010} showed that the last three methods correspond to a neighbourhood of functions (i.e. a tolerance tube)
defined by a specific metric for each method. Their topological approach constituted a unified framework named
virtual double-sided image probing (VDIP). The metrics were defined in the equivalence class of functions with a grey level addition of a constant. Only the patterns which were in the tolerance tube were selected. In practice, the tolerance tube was computed as a difference between a grey-scale dilation and an erosion. However, even if the approach of Barat et al. was based on a double-sided probing, such a method was not insensitive to lighting variations. It simply removed the variations due to an addition of a constant which had no optical justification contrary to the LIP functional Asplund's metrics. The existence of lighting variations in numerous settings such as medical images \cite{Noyel2014}, industrial control \cite{Noyel2011,Noyel2013}, driving assistance \cite{Hautiere2006a}, large databases \cite{Noyel2017c} or security \cite{Foresti2005} gives a prime importance to the functional Asplund's metrics defined in the LIP framework.

The aim of this paper is to study the existence of a link between the two functional Asplund's metrics. The paper is organised as follows. Firstly, the LIP framework will be reviewed. The definition and the properties of the LIP-multiplicative and the LIP-additive Asplund's metrics will then be recalled. Secondly, a link between the two metrics will be demonstrated. Finally, the results will be illustrated before concluding.


%
%
\section{Background}
\label{sec:back}

In this section, we will present the LIP model and the two functional Asplund's metrics.

\subsection{Logarithmic Image Processing}
\label{ssec:back:LIP}

The LIP model was introduced by Jourlin et al. \cite{Jourlin1988,Jourlin2001,Jourlin2016}. It is based on a famous optical law, namely the Transmittance law, which gives it nice optical properties, especially for processing low-light images. The LIP model is also consistent with the \textit{human visual system} as shown in \cite{Brailean1991}. This gives to that model the important property to process images acquired by reflection as a human eye would do. Let $\I = [0,M[^D$ be the set of grey level images defined on a domain $D \subset \Real^n$ with values in $[0,M[ \subset \Real$. For 8-bit digitised images, $M$ is equal to $2^8=256$. For an image $f \in \I$ acquired by transmission, the transmittance $T_f$ of the semi-transparent object which generates $f$ is equal to $T_f = 1 - f/M$. According to the transmittance law, the transmittance $T_{f \LP g}$ of the superimposition of two objects which generate $f$ and $g$ is equal to the (point-wise) product ``$.$'' of their transmittances $T_f$ and $T_g$:
\begin{equation}
	T_{f \LP g} = T_f . T_g. \label{eq:transLaw}
\end{equation}
By replacing the transmittances by their expressions in equation \ref{eq:transLaw}, one can deduce the LIP-addition of two images $f \LP g$:
\begin{equation}
	f \LIPplus g = f + g - fg/M. \label{eq:LIP:plus}%
\end{equation}
The LIP-multiplication $\LT$ of an image $f$ by a real number $\la$ is deduced from equation \ref{eq:LIP:plus} by considering that the addition $f \LP f$ may be written as $2 \LT f$:
\begin{equation}
	\la \LIPtimes f = M - M \left( 1 - f/M \right)^{\la}. \label{eq:LIP:times}%
\end{equation}
A LIP-negative function $\LM f$ can be defined by the equality $f \LP (\LM f) = 0$ which allows to write the LIP-difference $f \LM g$ between two images $f$ and $g$:
\begin{align}
	\LM f 	&= (-f)/(1-f/M) 	\label{eq:LIP:neg}\\
	f \LM g &= (f-g)/(1-g/M). \label{eq:LIP:minus}%
\end{align}
\begin{remark}
	$\LM f$ is not always an image, as it may have negative values. $f \LM g$ is an image iff $f \geq g$. As $\LM f$ may take values in the interval $]-\infty,M[$, it is called a function. The set of functions of which the values are less than $M$ is denoted $\Fcurv_M$ and is equal to $]-\infty,M[^D$.
\end{remark}

\begin{remark}
Contrary to the classical grey scale, the LIP-scale is inverted: $0$ corresponds to the white extremity when no obstacle is placed between the source and the sensor, whereas $M$ corresponds to the black extremity when no light passes through the object.
\end{remark}

\subsection{The LIP-multiplicative Asplund's metric}
\label{ssec:back:LIPMultAsp}


The LIP-multiplicative Asplund's metric was defined in \cite{Jourlin2012,Jourlin2014}. 
Let $\I^*={]0,M[}^D$ be the space of images with strictly positive values. 

\begin{definition}[LIP-multiplicative Asplund's metric]
Let $f$ and $g \in \I^*$ be two grey level images. A probing image is selected, e.g. $g$, and two numbers are defined: $\lambda = \inf \left\{\alpha, f \leq \alpha \LIPtimes g \right\}$ and $\mu = \sup \left\{\alpha, \alpha \LIPtimes g \leq f\right\}$. 
The LIP-multiplicative Asplund's metric $d_{As}^{\LIPtimes}$ is defined by:
\begin{eqnarray}
	d_{As}^{\LIPtimes}(f,g) &=& \ln \left( \lambda / \mu \right). \label{eq:dasMult}%
\end{eqnarray}
\label{def:dasMult}
\end{definition}

\begin{property}[\cite{Jourlin2012}]
The LIP-multiplicative Asplund's metric is theoretically invariant under lighting changes caused by variations of the semi-transparent object opacity (or thickness) which are modelled by a LIP-multiplication:
$\forall \alpha \in \Real^{*+}$: $d_{As}^{\LT}(f,g) = d_{As}^{\LT}(\alpha \LT f,g)$.
\end{property}

To be mathematically rigorous, $d_{As}^{\LIPtimes}$ is a metric in the space of equivalence classes $\I^{\LIPtimes}$ of the images $f^{\LT}$ and $g^{\LT}$, where $f^{\LIPtimes}=\{h \in \I / \exists k >0,  k \LIPtimes f = h  \}$ \cite[chap 3]{Jourlin2016}. However, one can keep the notations we used because
$\forall (f^{\protect \LIPtimes},g^{\protect \LIPtimes}) \in (\I^{\protect \LIPtimes})^2 , \quad d_{As}^{\protect \LIPtimes}(f^{\protect \LIPtimes},g^{\protect \LIPtimes}) = d_{As}^{\protect \LIPtimes}(f_1,g_1)$,
where $d_{As}^{\protect \LIPtimes}(f_1,g_1)$ is the Asplund's distance between any elements $f_1$ and $g_1$ of the equivalence classes $f^{\protect \LIPtimes}$ and $g^{\protect \LIPtimes}$. The relation $(\exists k >0,  k \LIPtimes f = h)$ is an equivalence relation which satisfies the three properties of reflexivity, symmetry and transitivity \cite{Jourlin2016,Noyel2017a}

\begin{remark}[Terminology]
When the Asplund's metric $d_{As}^{\LIPtimes}$ is applied between two images $f$ and $g$, the real value $d_{As}^{\LIPtimes}(f,g)$ is called the Asplund's distance between $f$ and $g$. The distance is the value of the metric between both images.
\end{remark}

Let $b \in ]0,M[^{D_b}$ be a probe function defined on a domain $D_b \subset D$. A map of Asplund's distances between an image $f \in \I^*$ and a probe $b$ can be introduced as follows for each pixel $x \in D$ of the image $f$.

\begin{definition}[Map of LIP-multiplicative Asplund's distances \cite{Jourlin2012}]
The map of Asplund's distances $As_{b}^{\LIPtimes}: \I^* \rightarrow \left(\Real^{+} \right)^{D}$ is defined by:
\begin{align}
As_{b}^{\LIPtimes} f(x) &= d_{As}^{\LIPtimes} (f_{\left|D_b(x)\right.},b).\label{eq:map_AsMult}%
\end{align}
\label{def:map_As_mult}
\end{definition}
$f_{\left|D_b(x)\right.}$ is the restriction of $f$ to the neighbourhood $D_b(x)$ centred on $x \in D$.

The map of the least upper bounds (mlub) $\la_b$ and the map of the greatest lower bounds (mglb) $\mu_b$ can also be defined as follows.
Let $\Ib = [0,M]^D$ be the set of images with the value $M$ included. 

\begin{definition}[LIP-multiplicative maps of the least upper and of the greatest lower bounds \cite{Noyel2017a}]
Given $\overline{\Real}^+=[0,+\infty]$, let $f \in \overline{\I}$ be an image and $b \in ]0,M[^{D_{b}}$ a probe. Their map of the least upper bounds (mlub) $\la_b: \overline{\I} \rightarrow (\overline{\Real}^{+} )^{D}$ and their map of the greatest lower bounds (mglb) $\mu_{b}: \overline{\I} \rightarrow (\overline{\Real}^{+} )^{D}$ are respectively defined by:
\begin{align}
		\la_{b} f(x) &=  \inf_{h \in D_b}{ \{ \alpha, f(x+h) \leq \alpha \LIPtimes b(h) \}}, \label{eq:upper_map}\\
		\mu_{b} f(x) &=  \sup_{h \in D_b}{ \{ \alpha, \alpha \LIPtimes b(h) \leq f(x+h) \}}. \label{eq:lower_map}%
	\end{align}
\end{definition}

The map of LIP-multiplicative Asplund's distance has also the property to be theoretically insensitive to lighting changes caused by variations of the object opacity (or thickness) which are modelled by a LIP-multiplication.

Let us define $\widetilde{f}= \ln{\left( 1 - f/M \right)}$, where $f \in \Ib$. The relation between the map of LIP-multiplicative Asplund's distances and MM has been demonstrated in \cite{Noyel2017a} with the following propositions, \ref{prop:genexpr_lowerUpperAspMaps} and \ref{prop:property_lower_upper_maps}.

\begin{proposition}
	Given $f \in \overline{\I}$, the mlub $\la_{b}$ and mglb $\mu_{b}$ are equal to:
	\begin{align}
	\la_{b} f(x) &= \vee{ \{ \widetilde{f}(x+h) / \widetilde{b}(h) , h \in D_b \} },   \label{eq:upper_map_2}\\
	\mu_{b} f(x) &= \wedge{ \{ \widetilde{f}(x+h) / \widetilde{b}(h) , h \in D_b \} }. \label{eq:lower_map_2}%
	\end{align}
	If in addition $f > 0$, the map of Asplund's distances expression $As_{b}^{\LIPtimes}$ becomes:
	\begin{align}
	As_{b}^{\LIPtimes}f &= \ln \left( \la_{b} f / \mu_{b} f \right).\label{eq:map_AsMult_la_mu_general_se}%
	\end{align}
	\label{prop:genexpr_lowerUpperAspMaps}
\end{proposition}

\begin{proposition}
	The mlub $\la_b$ and the mglb $\mu_b$ are a dilation and an erosion, respectively, between the two complete lattices $(\overline{\I} , \leq)$ and $((\overline{\Real}^{+})^{D},\leq)$ \cite{Serra1982,Heijmans1994}.
	\label{prop:property_lower_upper_maps}
\end{proposition}
Indeed, $\forall f,g \in \Ib$, the mlub distributes over supremum $\la_b(f \vee g) = \la_b(f) \vee \la_b(g)$ and the mglb distributes over infimum $\mu_b(f \wedge g) = \mu_b(f) \wedge \mu_b(g)$ \cite{Noyel2017a}. 

Moreover, as shown in \cite{Noyel2017b}, the mlub, mglb and distance map will be expressed in proposition \ref{prop:lower_upper_AsDist_maps_MM_op} with the usual morphological operations for grey level functions, namely the dilation $(f \oplus b)(x) = \vee_{h\in D_b}\{f(x-h)+b(h)\}$ and the erosion $(f \ominus b)(x) = \wedge_{h\in D_b}\{f(x+h)-b(h)\}$.
The symbols $\oplus$ and $\ominus$ represent the extension to functions of the Minkowski operators between sets.

\begin{proposition}
	The mlub $\la_b$, the mglb $\mu_b$ and the distance map $As_b^{\LT}$ are equal to \cite{Noyel2017b}:
	\begin{align}
		\la_{b} f &= \exp{ ( \widehat{f} \oplus (- \widehat{\overline{b}}) ) }, \label{eq:upper_map_morpho}\\
		\mu_{b} f &= \exp{ ( \widehat{f} \ominus \widehat{b} ) }, \label{eq:lower_map_morpho}\\
		As_{b}^{\LIPtimes}f &= \left[ \widehat{f} \oplus (- \widehat{\overline{b}}) \right] - \left[ \widehat{f} \ominus \widehat{b} \right].\label{eq:As_mapMult_morpho}%
	\end{align}
	$\overline{b}$ is the reflected structuring function defined by $\forall x \in \overline{D}_b$, $\overline{b}(x)=b(-x)$ \cite{Soille2003} and $\widehat{f}$ is the function defined by $\widehat{f} = \ln{(- \widetilde{f})} = \ln{(- \ln{\left( 1 - f/M \right)})}$.
	\label{prop:lower_upper_AsDist_maps_MM_op}
\end{proposition}

As the operations of dilation $\oplus$ and erosion $\ominus$ exist in numerous image analysis software, equation \ref{eq:As_mapMult_morpho} facilitates the programming of the map of LIP-multiplicative Asplund's distances $As_{b}^{\LIPtimes}f$ of the image $f$.

\subsection{The LIP-additive Asplund's metric}
\label{ssec:back:LIPAddAsp}

Jourlin has proposed a definition for the LIP-additive Asplund's metric \cite[chap. 3]{Jourlin2016}. Let us present it in a more precise way. 

\begin{definition}[LIP-additive Asplund's metric]
Let $f$ and $g \in \Fcurv_M $ be two functions, we select a probing function, e.g. $g$, and we define the two
numbers: $c_1 = \inf{ \{c, f \leq c \LP g \}}$ and $c_2 = \sup{ \{c, c \LP g \leq f \} }$, where $c$ lies in the interval $]-\infty,M[$.
The LIP-additive Asplund's metric $d_{As}^{\protect \LP}$ is defined according to:
\begin{eqnarray}
	d_{As}^{\LP}(f,g) &=& c_1 \LM c_2. \label{eq:dasAdd}%
\end{eqnarray}
\label{def:dasAdd}
\end{definition}

\begin{property}[{\cite[chap. 3]{Jourlin2016}}]
The LIP-additive Asplund's metric is theoretically invariant under lighting changes caused by variations of the light intensity (or camera exposure-time) which are modelled by a LIP-addition of a constant:
$\forall k \in ]-\infty,M[$, $d_{As}^{\LP}(f,g) = d_{As}^{\LP}(f \LP k , g)$ and $d_{As}^{\LP}(f,f \LP k) = 0$.
\end{property}

\begin{remark}
As for the LIP-multiplicative Asplund's metric, it would be more rigorous to define the LIP-additive Asplund's metric on the equivalence classes $f^{\LP}$ and $g^{\LP}$, where $f^{\LP}=\{h \in \Fcurv_M / \exists k \in ]-\infty,M[,  f \LP k = h  \}$.
\end{remark}

A map of Asplund's distances between an image and a probe can also be defined for each image pixel.

\begin{definition}[Map of LIP-additive Asplund's distances]
Let $f \in \Fcurv_M$ be a function and $b \in ]-\infty,M[^{D_{b}}$ a probe. The map of Asplund's distances is the mapping $As_{b}^{\LIPplus}: \Fcurv_M \rightarrow \I$ defined by:
\begin{align}
	As_{b}^{\LIPplus} f(x) &= d_{As}^{\LIPplus} (f_{\left|D_b(x)\right.},b). \label{eq:map_As_add}%
\end{align}
\label{def:map_As_add}
\end{definition}

The map of LIP-additive Asplund's distance has also the property to be theoretically insensitive to lighting changes caused by variations of light intensity (or camera exposure-time) which are modelled by a LIP-addition of a constant.

%
%
\section{Linking the LIP-multiplicative and the LIP-additive Asplund's metrics}
\label{sec:link}

In this section, first, the map of LIP-additive Asplund's distances will be expressed with neighbourhood operations. Then the link between both maps and the metrics will be studied. Finally, this link will be briefly discussed.

\subsection{General expression of the map of LIP-additive Asplund's distances}
\label{ssec:link:LIPAddAsp}

From definition \ref{def:dasAdd}, the maps of the least upper bounds $c_{1_{b}} f$ and of the greatest lower bounds $c_{2_{b}} f$ can be defined by computing the constant $c_1$ and $c_2$ between a probe $b \in  ]-\infty,M[^{D_b}$ and the function restriction $f_{|D_b(x)}$.

\begin{definition}[LIP-additive maps of the least upper and of the greatest lower bounds]
Let $f \in \Fcurvb_M$ be a function and $b \in ]-\infty,M[^{D_{b}}$ a probe. Their map of the least upper bounds (mlub) $c_{1_b}: \Fcurvb_M  \rightarrow \Fcurvb_M$ and their map of the greatest lower bounds (mglb) $c_{2_{b}}: \Fcurvb_M \rightarrow \Fcurvb_M$ are defined by:
\begin{align}
	c_{1_{b}} f(x) &=  \inf_{h \in D_b}{ \{c, f(x+h) \leq c \LIPplus b(h) \} }  \label{eq:upper_map_add}\\
	c_{2_{b}} f(x) &=  \sup_{h \in D_b}{ \{c, c \LIPplus b(h)	\leq f(x+h) \} }.\label{eq:lower_map_add}%
\end{align}
\end{definition}

The mlub expression can be rewritten as follows, $\forall x \in D$:
\begin{align}
c_{1_{b}} f(x) &= \inf \{c, c \geq f(x+h) \LIPminus b(h), h \in D_b \}\nonumber\\
						&= \vee \{ f(x+h) \LIPminus b(h) , h \in D_b \}, \label{eq:upper_map_add_2}
\end{align}	
where the last equality is due to the complete lattice structure. In a similar way, the mglb $c_{2_{b}}$ becomes:
\begin{align}
c_{2_{b}} f(x) 
						&= \wedge \{ f(x+h) \LIPminus b(h), h \in D_b \}.\label{eq:lower_map_add_2}
\end{align}
The general expression of the map of LIP-additive Asplund's distances between $f$ and $b$ is therefore:
\begin{align}
	As_{b}^{\LIPplus}f &= c_{1_{b}} f \LIPminus c_{2_{b}} f. \label{eq:map_As_c1_c2_add}%
\end{align}
This last expression will be useful to establish the link with the map of LIP-additive Asplund's distances.

\subsection{Link between the maps of distances (and the metrics)}
\label{ssec:link:link}

First of all, an isomorphism is needed between the lattice $(\Realb^{+})^D$ of the LIP-multiplicative mlub $\la_b f$, or mglb $\mu_b f$, and the lattice $[-\infty,M]^D$ of the LIP-additive mlub $c_{1_b} f$, or mglb $c_{2_b} f$. This isomorphism $\xi: [-\infty,M]^D \rightarrow \Realb^D$ and its inverse $\xi^{-1}$ were both defined in \cite{Jourlin1995,Navarro2013} by:
\begin{align}
	\xi(f) 			&= 	-M \ln{(1-f/M)} 		\label{eq:isomorph_inv_LIP_add}\\
	\xi^{-1}(f) &= 	M(1-\exp{(-f/M)}). \label{eq:isomorph_LIP_add}%
\end{align}

\begin{remark}
One can notice that $\xi(f) = -M \tilde{f}$. This relation will be useful in the proof hereinafter.
\end{remark} 
There exists the following relation between the distance maps.


\begin{proposition}
	Let  $f \in \I^*$ be an image, $f^c=M-f$ its complement and $b \in ]0,M[^{D_b}$ a structuring function. The map of LIP-additive Asplund's distances is related to the map of LIP-multiplicative distances by the following equation:
	\begin{align}
		As_b^{\LT} f &= \frac{1}{M} \xi \left( As_{\left(\xi(b)\right)^c}^{\LIPplus}\left(\xi(f)\right)^c \right), \label{eq:rel_As_Add_Mult}
	\end{align}
	where $(\xi(f))^c = M - \xi(f) \in ]-\infty,M[^D$.
	\label{prop:rel_As_Add_Mult}
\end{proposition}

\begin{corollary}
	Using the following variable changes $f_1 = \left(\xi(f)\right)^c$ and $b_1 = \left(\xi(b)\right)^c$, with $f_1 \in \Fcurv_M$ and $b_1 \in ]-\infty,M[^{D_{b}}$, an equivalent equation is obtained:
\begin{align}
As_{b_1}^{\LP} f_1 &= \xi^{-1} \left(M. As_{\xi^{-1}(b_1^c)}^{\LT} \xi^{-1}(f_1^c) \right). \label{eq:rel_As_Mult_Add}%
\end{align}
\label{prop:rel_As_Mult_Add}
\end{corollary}

\begin{remark} Equation \ref{eq:rel_As_Add_Mult} can also be written as: \\$As_{(\xi(b))^c}^{\LP}(\xi(f))^c =  \xi^{-1}( M. As_b^{\LT} f ) = M ( 1 - \exp{(-As_b^{\LT} f)} )$.
As the map of LIP-multiplicative Asplund's distances $As_b^{\LT} f$ of $f$ is an element of $[0,+\infty[^D$, the map of LIP-additive distances $As_{(\xi(b))^c}^{\LP}(\xi(f))^c$ of $(\xi(f))^c$ is an element of $[0,M[^D=\I$ and is therefore an image.
\end{remark}

\begin{remark}
The same relation exists between both functional Asplund's metrics:
\begin{equation}
d_{As}^{\LT} (f,g) = \frac{1}{M}\xi( d_{As}^{\LIPplus} ( [\xi(f)]^c , [\xi(g)]^c )).  \label{eq:rel_metricAs_Add_Mult}
\end{equation}
\end{remark}


\begin{proof}
Let $f \in \Ib = [0,M]^D$ be an image, there is: 
\begin{align}
		(M - f) \LIPminus (M - b)	&= M\frac{(M - f) - (M - b)}{M - (M - b)}=M\frac{b - f}{b} = M \left(1 - \frac{f}{b} \right).\label{eq:LIP_minus_compt}%
	\end{align}

Equation \ref{eq:LIP_minus_compt} is set in equations \ref{eq:upper_map_add_2} and \ref{eq:lower_map_add_2}, which gives: $\forall x \in D$, 
	\begin{align}
	  c_{1_{(\xi(b))^c}}^{\LIPplus}\left(\xi(f)\right)^c(x) 
		&= \vee_{h \in D_b} \{ (M-\xi(f)(x+h)) \LIPminus (M-\xi(b)(h)) \} \nonumber\\
		&= \vee_{h \in D_b} \left\{ M \left(1 - \frac{\xi(f)(x+h)}{\xi(b)(h)} \right) \right\} \nonumber\\
		&= M  \left(1 - \wedge_{h \in D_b} \left\{ \frac{\xi(f)(x+h)}{\xi(b)(h)} \right\} \right) \nonumber\\
		&= M  \left(1 - \wedge_{h \in D_b} \left\{ \frac{-M\tilde{f}(x+h)}{-M\tilde{b}(h)} \right\} \right) \nonumber\\
		&= M  (1 - \wedge_{h \in D_b} \{ \tilde{f}(x+h) / \tilde{b}(h) \} ) \nonumber\\
		&= M  (1 - \mu_b f(x) ), \label{eq:upper_map_compt2}\\
		c_{2_{(\xi(b))^c}}^{\LIPplus}\left(\xi(f)\right)^c(x) &=  M(1 - \vee_{h \in D_b} \{\tilde{f}(x+h)/ \tilde{b}(h) \}) \nonumber\\
		&= M  (1 - \la_b f(x) ). \label{eq:lower_map_compt2}
	\end{align}
By combining equations \ref{eq:upper_map_compt2} and \ref{eq:lower_map_compt2} with equations \ref{eq:LIP_minus_compt}, \ref{eq:map_AsMult_la_mu_general_se} and \ref{eq:isomorph_LIP_add}, one deduces the following equations:
	\begin{align}
		As_{(\xi(b))^c}^{\LIPplus}\left(\xi(f)\right)^c &=  c_{1_{(\xi(b))^c}}^{\LIPplus}\left(\xi(f)\right)^c \LM c_{2_{(\xi(b))^c}}^{\LIPplus}\left(\xi(f)\right)^c 
		= (M - M\mu_b f ) \LM (M - M\la_b f )\nonumber \\
		&= M \left( 1 -\frac{ \mu_b f }{ \la_b f } \right) 
		= M ( 1 - \exp{(-As_b^{\LT} f)} ) \nonumber\\
		&= \xi^{-1} (M . As_b^{\LT} f) \nonumber\\
		\Leftrightarrow As_b^{\LT} f &= \frac{1}{M} \xi \left( As_{\left(\xi(b)\right)^c}^{\LP}\left( \xi(f) \right)^c \right). \nonumber
	\end{align}	
	
Moreover, when $f$ lies in $\Ib$, the function $\xi(f)$ lies in $[0,\infty]^D$ and
$(\xi(f))^c = M -\xi(f)$ lies in $[-\infty,M]^D$.
\qed
\end{proof}

\subsection{Discussion}
\label{ssec:link:disc}

Equations \ref{eq:rel_As_Mult_Add} and \ref{eq:rel_As_Add_Mult} show that the maps of LIP-multiplicative and LIP-additive Asplund's distances as well as their corresponding metrics are related by the isomorphism $\xi$. These relations allow to compute one distance map of an image by mean of the other distance map of a transform of this image. E.g. the LIP-additive map $As_{b_1}^{\LP} f_1$ of the function $f_1$ can be computed by using the program of the LIP-multiplicative map $As_{\xi^{-1}(b_1^c)}^{\LP} \xi^{-1}(f_1^c)$ of the transformed function $\xi^{-1}(f_1^c)$. However, both equations do not directly link both distance maps of the image $f$. E.g. the LIP-multiplicative map of the image $f$, $As_b^{\LT} f$, is not directly related to the LIP-additive map of the image $f$, $As_b^{\LP} f$. 

The relation given in equations \ref{eq:rel_As_Mult_Add} and \ref{eq:rel_As_Add_Mult} is not surprising. Indeed, the map of LIP-additive Asplund's distances $As_{\left(\xi(b)\right)^c}^{\LIPplus}\left(\xi(f)\right)^c$ is an image which lies in $[0,M[^D$, whereas the map of LIP-multiplicative Asplund's distances $As_b^{\LT} f$ lies in $(\Real^+)^D$. The isomorphism $\xi$ allows to pass from the image space of the LIP-additive distance map $As_{\left(\xi(b)\right)^c}^{\LIPplus}\left(\xi(f)\right)^c$ to the real function space of the LIP-multiplicative distance map $As_b^{\LT} f$.

%
%
\section{Illustration}
\label{sec:ill}

\begin{figure}[!htb]
\centering
\subfloat[]{\includegraphics[width=0.31\columnwidth]{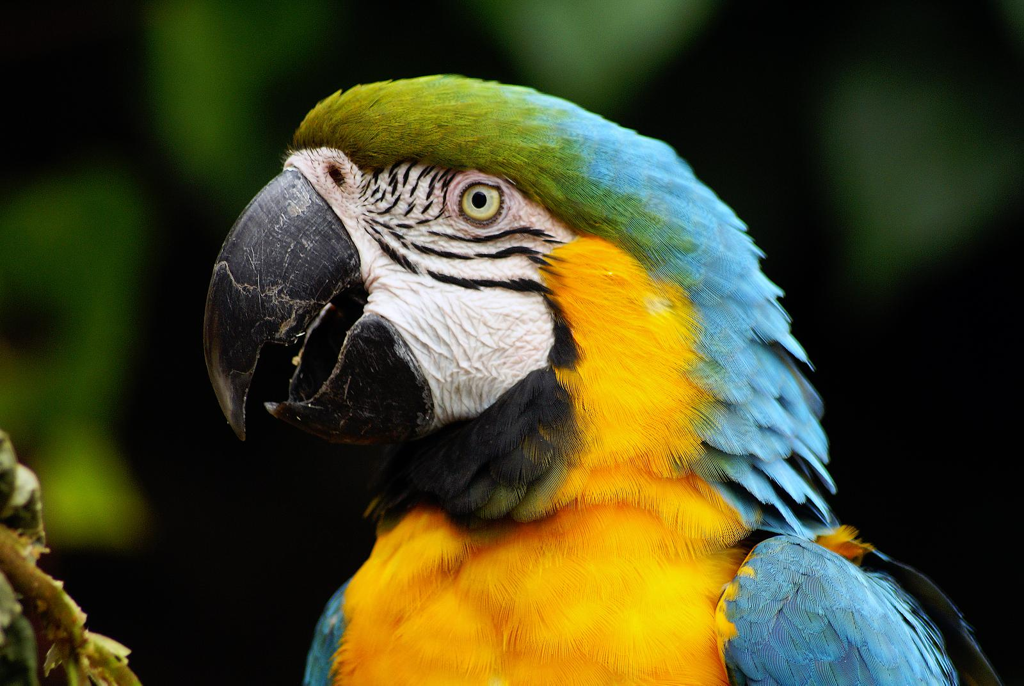}%
\label{fig:Parot:rel_map_dist_add:1}}
\hfil
\subfloat[]{\includegraphics[width=0.31\columnwidth]{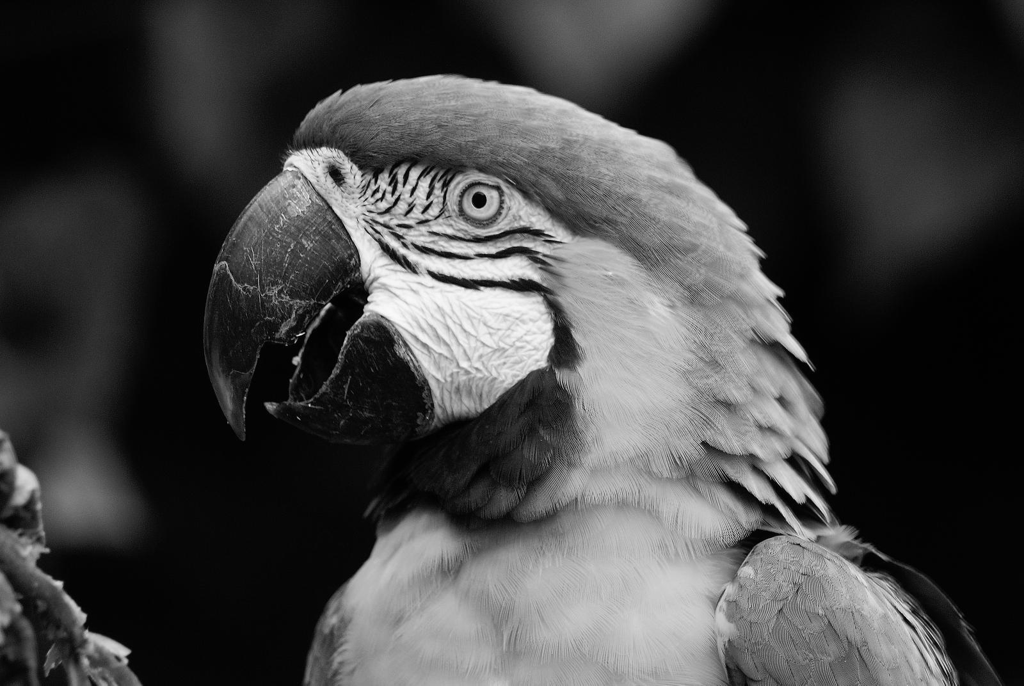}%
\label{fig:Parot:rel_map_dist_add:2}}
\hfil
\subfloat[]{\includegraphics[width=0.31\columnwidth]{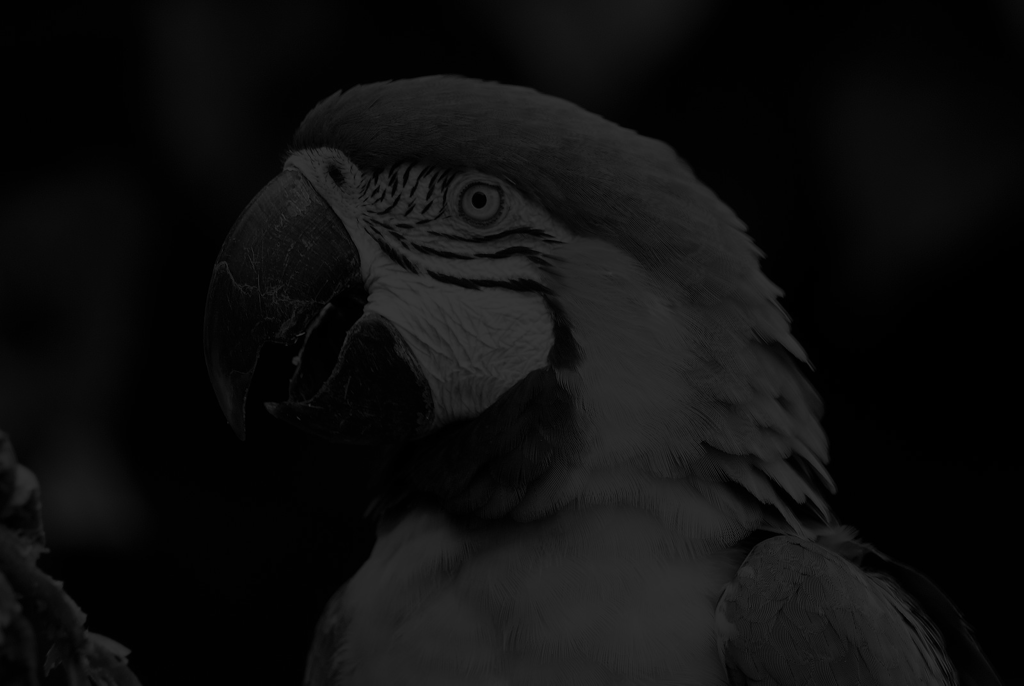}%
\label{fig:Parot:rel_map_dist_add:3}}
\hfil
\subfloat[]{\includegraphics[width=0.25\columnwidth]{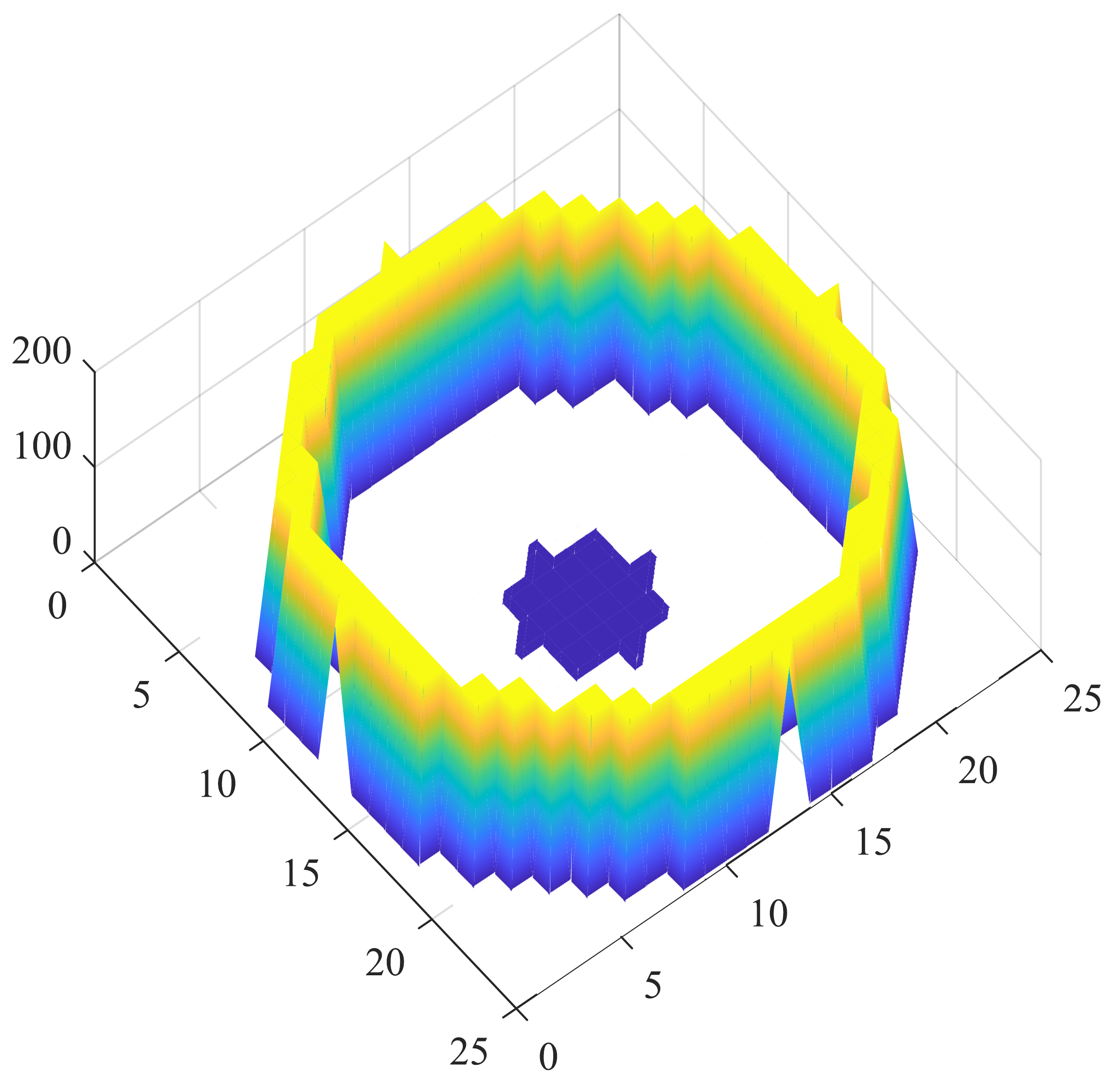}%
\label{fig:Parot:rel_map_dist_add:4}}\\
\subfloat[]{\includegraphics[width=0.31\columnwidth]{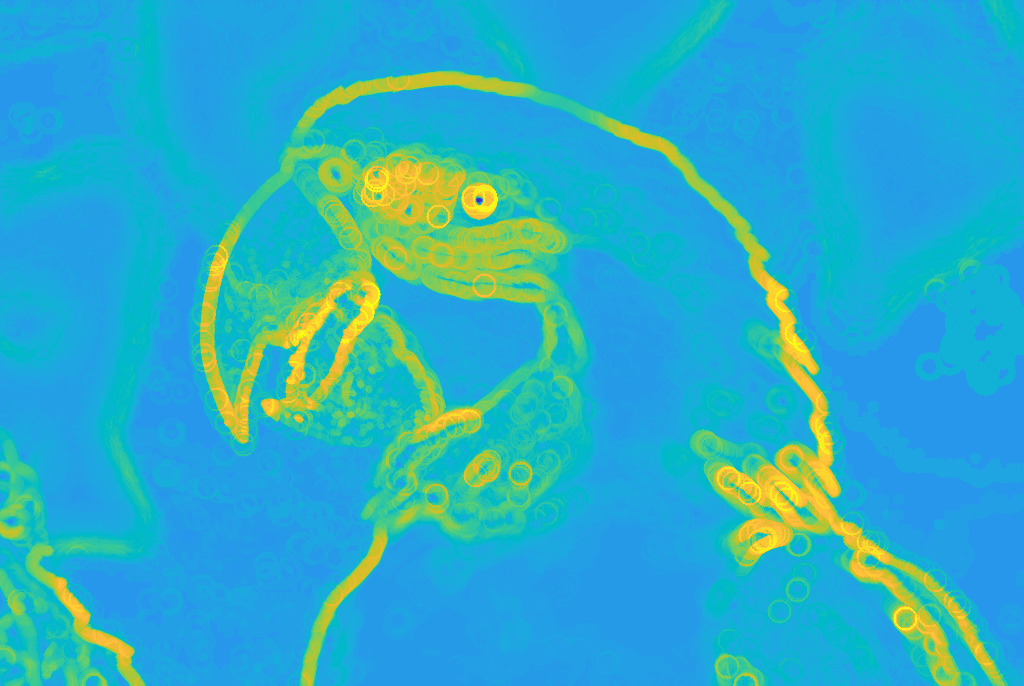}%
\label{fig:Parot:rel_map_dist_add:5}}
\hfil
\subfloat[]{\includegraphics[width=0.31\columnwidth]{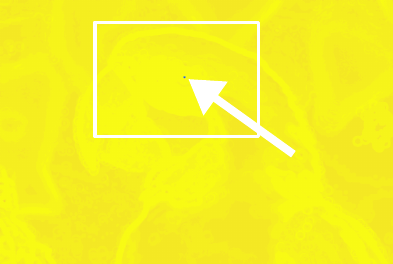}%
\label{fig:Parot:rel_map_dist_add:6}}
\hfil
\subfloat[]{\includegraphics[width=0.31\columnwidth]{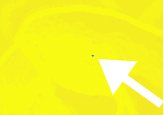}%
\label{fig:Parot:rel_map_dist_add:7}}
\caption{(a) Colour image of a parrot \cite{YFCC100Mparrot2008} and (b) its luminance $f$. (c) Darkened image $f^{dk}$ obtained by a LIP addition of a constant 200: $f^{dk} = f \protect \LP 200$. (d) Topographic surface of the probe $b$. (e) Map of LIP-multiplicative Asplund's distances $As_{\xi^{-1}(b^c)}^{\protect \LT} \xi^{-1}(f^c)$ of the function $\xi^{-1}(f^c)$ which is used to compute the (f) map of LIP-additive Asplund's distances $As_{b}^{\protect \LP} f$ of the image $f$. It is also equal to the distance map $As_{b}^{\protect \LP} f^{dk}$ between the darkened image $f^{dk}$ and the same probe $b$. (g) Zoom in of the map (f). Both white arrows indicate the minimum of the map $As_{b}^{\protect \LP} f$ corresponding to the eye centre.}
\label{fig:Parot:rel_map_dist_add}
\end{figure}

Figure \ref{fig:Parot:rel_map_dist_add} illustrates relation \ref{eq:rel_As_Add_Mult}, where the map of LIP-additive Asplund's distances $As_{b}^{\LP}$ is deduced from the map of LIP-multiplicative distances $As_{\xi^{-1}(b^c)}^{\LT}$. Moreover, it shows the theoretical insensitivity of the map of LIP-additive Asplund's distances to a lighting change simulated by the LIP-addition of a constant. 
For this experiment, an image of a parrot \cite{YFCC100Mparrot2008} (Fig. \ref{fig:Parot:rel_map_dist_add:1}) was selected in the Yahoo Flickr Creative Commons 100 Million Dataset (YFCC100M) \cite{Thomee2016} and converted into a luminance image $f$ in grey levels (Fig. \ref{fig:Parot:rel_map_dist_add:2}). A darkened image $f^{dk}$ is obtained by LIP-adding a constant 200 to the luminance image $f$: $f^{dk} = f \LP 200$ (Fig. \ref{fig:Parot:rel_map_dist_add:3}). This operation simulates a decreasing of the camera exposure-time or a decreasing of the light intensity. In order to detect the parrot's eye, a probe $b$ is designed. A white ring - with a height of 161 grey levels - surrounds a black disk whose grey value is equal to 4 (Fig \ref{fig:Parot:rel_map_dist_add:4}). The LIP-additive distance map $As_{b}^{\LP} {f}$ of the image $f$ (Fig. \ref{fig:Parot:rel_map_dist_add:6} and \ref{fig:Parot:rel_map_dist_add:7}) is computed from the LIP-multiplicative distance map $As_{\xi^{-1}(b^c)}^{\LT} \xi^{-1}(f^c)$ of the function $\xi^{-1}(f^c)$ (Fig. \ref{fig:Parot:rel_map_dist_add:5}) using equation \ref{eq:rel_As_Add_Mult}. The centre of the parrot's eye corresponds to the minimum of this former map, $As_{b}^{\LP} {f}$ (Fig. \ref{fig:Parot:rel_map_dist_add:6} and \ref{fig:Parot:rel_map_dist_add:7}), which can be easily extracted by a threshold. It is remarkable that the detection of an object in a low-light and complex image can be performed by a simple threshold of its map of Asplund's distances.

Moreover, the LIP-additive distance map $As_{b}^{\LP} {f^{dk}}$ between the darkened image $f^{dk}$ and the probe $b$ - designed for the brightest image $f$ - is also equal to the LIP-additive distance map $As_{b}^{\LP} {f}$ between the brightest image $f$ and the probe $b$ (Fig. \ref{fig:Parot:rel_map_dist_add:6}). This result shows the insensitivity of the map of LIP-additive Asplund's distance to a variation of camera exposure-time which is simulated by a LIP-addition of a constant.

In addition, the LIP-additive distance map was also computed directly with equation \ref{eq:map_As_add} and compared to the LIP-additive distance map obtained from equation \ref{eq:rel_As_Add_Mult} (Fig. $\ref{fig:Parot:rel_map_dist_add:6}$). Both maps of distances were equal with a numerical precision corresponding to the rounding error of the computer. 
Therefore, relation \ref{eq:rel_As_Add_Mult} is numerically verified.

%
%
\section{Conclusion}
\label{sec:concl}

A link between the maps of LIP-multiplicative and LIP-additive Asplund's distances has therefore been successfully demonstrated. The relation is based on the LIP-isomorphism which allows to pass from the image space of the LIP-additive distance map to the positive real function space of the LIP-multiplicative distance map. However, there does not exist a link between the maps of LIP-multiplicative and LIP-additive Asplund's distances of the same image. Nevertheless, the proven relation is not only interesting from a theoretical point of view but also from a practical point of view. Indeed, it allows to compute one map of an image from the other map of a transform of this image.  Experiments have verified the relation from a numerical point of view. They have also illustrated the main interest of the map of LIP-additive Asplund's distances, i.e. its insensitivity to lighting changes modelled by a LIP-addition and corresponding to a variation of the camera exposure-time or of the light intensity. Such properties open the way to numerous applications where the lighting conditions are partially controlled.



\bibliographystyle{splncs04}
\bibliography{refs}

\end{document}